\documentclass[runningheads]{llncs}
\usepackage{graphicx}
\usepackage{cite}
\usepackage{tikz}
\usepackage{comment}
\usepackage{amsmath,amssymb} 
\usepackage{color}

\usepackage{amsmath}
\usepackage{subfigure}
\usepackage{booktabs}
\usepackage{amssymb}
\usepackage{mathtools}

\DeclarePairedDelimiter{\ceil}{\lceil}{\rceil}

\usepackage{algorithm}
\usepackage{algorithmic}
\usepackage{setspace}
\usepackage{array}
\usepackage{pifont}
\usepackage{wrapfig}
\usepackage{hyperref}
\usepackage[accsupp]{axessibility}  

\newcommand{\eg}{\textit{e}.\textit{g}.}

\newcommand{\cmark}{\ding{51}}%
\begin{document}
\pagestyle{headings}
\mainmatter
\def\ECCVSubNumber{3315}  

\title{Pyramid Region-based Slot Attention Network for Temporal Action Proposal Generation} 


\titlerunning{PRSA-Net}
%
\author{Shuaicheng Li$^{1}$, Feng Zhang$^{1}$, Rui-Wei Zhao$^{3}$, Rui Feng$^{2,3,4}$, Kunlin Yang$^{1}$,\\ Lingbo Liu$^{5}$, Jun Hou$^{1}$}
%
\authorrunning{Shuaicheng Li et al.}
%
\institute{Sensetime Research \and
Shanghai Key Laboratory of Intelligent Information Processing,\\
School of Computer Science, Fudan University, China \and
Academy for Engineering and Technology, Fudan University, China \and
Fudan Zhangjiang Institute, Shanghai \and
The Hong Kong Polytechnic University, China}
\maketitle

\begin{abstract}
It has been found that temporal action proposal generation, which aims to discover the temporal action instances within the range of the start and end frames in the untrimmed videos, can largely benefit from proper temporal and semantic context exploitation.
The latest efforts were dedicated to considering the temporal context and similarity-based semantic context through self-attention modules.
However, they still suffer from cluttered background information and limited contextual feature learning. 
In this paper, we propose a novel Pyramid Region-based Slot Attention (PRSlot) modules to address these issues.
Instead of using the similarity computation, our PRSlot module directly learns the local relations in an encoder-decoder manner and generates the representation of a local region enhanced based on the attention over input features called \textit{slot}.
Specifically, upon the input snippet-level features, PRSlot module takes the target snippet as \textit{query}, its surrounding region as \textit{key} and then generates slot representations for each \textit{query-key} slot by aggregating the local snippet context with a parallel pyramid strategy.
Based on PRSlot modules, we present a novel Pyramid Region-based Slot Attention Network termed PRSA-Net to learn a unified visual representation with rich temporal and semantic context for better proposal generation.
Extensive experiments are conducted on two widely adopted THUMOS14 and ActivityNet-1.3 benchmarks. 
Our PRSA-Net outperforms other state-of-the-art methods. In particular, we improve the AR@100 from the previous best 50.67\% to 56.12\% for proposal generation and raise the mAP under 0.5 tIoU from 51.9\% to 58.7\% for action detection on THUMOS14. \textit{Code is available at} \url{https://github.com/handhand123/PRSA-Net}

\end{abstract}

\section{Introduction}
\label{sec:intro}

Temporal action detection is a popular and fundamental problem for video content understanding.
This task aims to predict the precise temporal boundaries and categories of each action instance in the videos.
Similar to two-stage object detection in images, most temporal action detection methods \cite{Lin:2018:BSN,Lin:2019:BMN, Xu:2020:G-TAD,RTDNet:2021} follow a two-stage strategy: temporal action proposal generation and action proposal classification.
Action classification \cite{Wang:2017:Unet} has achieved convincing performance, but temporal action detection accuracy is still unsatisfactory on mainstream benchmarks.
It indicates that the quality of generated proposals is the bottleneck of the final action detection performance.
\begin{figure}[t]
\centering
\includegraphics[width=0.8\linewidth]{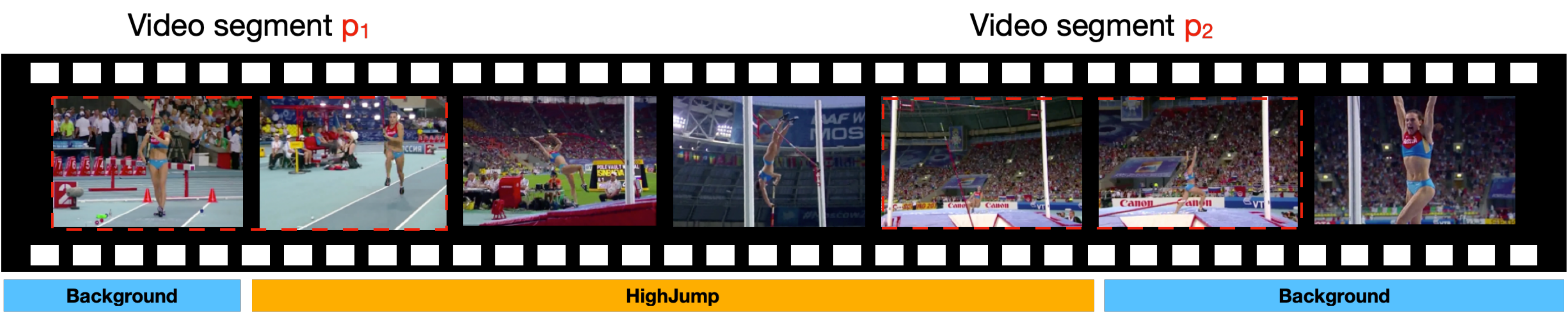}
\caption{
Given an untrimmed video sequence, we aim to generate temporal action instance.
The boundaries can hardly distinguished from background due to cluttered content and  action-unrelated scenes.
Our tailor-modified PRSA-Net is designed to selectively contextualize local semantic information instead of pairwise similarity.}
\label{fig:intro}
\end{figure}
Existing proposal generation approaches make sorts of efforts to exploit rich semantic information for the sake of better high-quality proposal generation.
The majority of previous works \cite{Lin:2018:BSN,Lin:2019:BMN,Lin:2020:Fast} embed the temporal representation by stacked temporal convolutions.
\cite{Chao:2018:TAL-Net,Zhao:2017:SSN} extend the temporal segment by a pre-defined ratio to capture contextual content, while G-TAD \cite{Xu:2020:G-TAD} proposes to explicitly model temporal contextual information by graph convolutional networks (GCNs).
More recently, like DETR \cite{DETR2020} in object detection, transformer based methods \cite{RTDNet:2021,activity:2021} are introduced to provide long-range temporal context for proposal generation and action detection.

Despite impressive progress, these aforementioned approaches still confront two challenges that remain to be addressed:
1) limited contextual representation learning and
2) sensitivity to the complicated background.
In the former case, although video snippets contain richer information than a single image such as temporal dynamical content, offering useful cues for generating proposals, much fewer efforts are spent on semantic temporal content modeling.
Therefore, effectively capturing action-matter context is critical for generating proposals.
In the latter case, due to the complicated background in untrimmed videos, \cite{Xu:2020:G-TAD} adopts a similarity-based attention mechanism to selectively extract the most relevant information and relationships.
However, it is sub-optimal to exploit the dependencies only based on computing the pairwise similarity.
Because there exists highly similar content for consistent frames, where the pairwise relations may introduce spurious noise such as cluttered background and inaccurate action-related content.
As illustrated in Fig. \ref{fig:intro}, the action instances surrounded by the background content, such as \textit{video segments} $p_1$, $p_2$, can hardly be discriminated due to the camera motion and progressive action transitions.
It demonstrates that applying traditional pairwise similarity-based relations such as the operation \textit{dot product, Euclidean distance} are insufficient to deliver complete contextual information for proposal generation.

To relieve the above issues, we propose a novel Pyramid Region-based Slot Attention (PRSlot) to contextualize action-matter representation. 
It is building upon the recently-emerged slot attention \cite{slotattn:2020}, which learns the object-centric representations for image classification task.
Our well-design PRSlot module takes the snippet-level features as input and maps them to a set of output vectors by aggregating local region context that we refer to as \textit{slots}. 
\textbf{First}, our PRSlot module is enhanced by a Region-based Attention (RA) which directly estimates the confidence from inputs and its local region to the slots.
Unlike the similarity-based attention mechanism, our RA restricts the scope of slot interactions to local surroundings and learns the semantic attention directly using an encoder-decoder architecture.
In detail, an encoder is deployed to exploit the snippet-relevant feature map in the local region by zeroing out the position outside of the desired local scope, while a decoder is followed to map the correlation features into the relation score vectors directly.
\textbf{Second}, instead of applying recurrent function to update slots over multiple iterations in original slot attention \cite{slotattn:2020}, our tailor-modified PRSlot module presents a parallel pyramid slot representation updating strategy with multi-scale local regions. 
\textbf{Finally}, the complementary action-matter representation generated by the PRSlot modules are used to produce boundaries scores and proposal-level confidence respectively by linear layers.
Based on the above components, we present a novel Pyramid Region-based Slot Attention Network called PRSA-Net to capture abundant semantic representation for high-quality proposal generations.
Experimental results show our PRSA-Net is superior to the state-of-the-art performance on two widely used benchmarks.
The main contributions of this paper are therefore as follows,
\begin{itemize}
\item A newly Region-based Attention is proposed to directly generate relation scores from the slot representations and its surroundings using encoder-decoder manner instead of similarity-based operation.
\item
We propose a novel Pyramid Region-based Slot Attention (PRSlot) module, which incorporates a region-based attention mechanism and a parallel pyramid iteration strategy to effectively capture contextual representations for better boundaries discrimination. 
Based on this module, we develop a Pyramid Region-based Slot Attention Network (PRSA-Net) to exploit action-matter information for high-quality proposal generation.


\item We perform extensive experiments on the THUMOS14 and ActivityNet-1.3 benchmarks and evaluate the performances of temporal action proposal generation and detection.
The results show that our proposed approach outperforms other state-of-the-art methods.
Our ablation study also highlights the importance of both architecture components.
\end{itemize}
\section{Related Work}
\label{sec:related}

\textbf{Video Feature Extraction.}
Since video data can be generally treated as a sequence of ordered frames, the most straightforward approach to video feature extraction is to first generate deep features on every frame by CNN backbones with 2D convolutional filters, and then fuse them into snippet or video level features \cite{Karpathy:2014:VidConv}.
Besides the 2D filters, 3D or (2+1)D convolutional filters powered backbones \cite{Tran:2015:3DConv,Carreira:2017:I3D,Qiu:2017:P3D} have been designed to directly extract features from multiple stacked video frames, which are more suitable for motion feature extraction.
Another widely used framework to explicitly incorporate motion clues is the two-stream CNNs \cite{Simonyan:2014:2S-CNN, Feichtenhofer:2016:2S-CNN}, in which one stream with 2D CNN is deployed on the RGB frames, while the other stream extracts features from the optical flows.
In this work, we adopt the Kinetics \cite{Kay:2017:Kinetics} pretrained inflated 3D network (I3D)\cite{Carreira:2017:I3D} as the feature extractor.
We consider multi-modal information such as RGB and optical flow information in the adopted backbone.
Also, for a fair comparison, we also conduct experiments using the TSN \cite{Wang:2016:TSN}, where ResNet \cite{resnet2016} network and BN-Inception \cite{inception2015} network are used as the spatial and temporal network respectively.

\noindent
\textbf{Temporal Action Proposals and Detection.}
Based on the extracted video features from the feature extractor, numbers of recent approaches solve the action detection problems in two steps:
(1) proposal generation;
(2) proposal classification and refinement.
In the first step, the proposals can be generated in a top-down fashion, which is based on preset sliding temporal windows or anchors \cite{Shou:2016:S-CNN, Heilbron:2016:TAP, Gao:2017:CBR, Chao:2018:TAL-Net}.
Or alternatively, the proposals can be generated in a bottom-up fashion, which directly predicts proposal locations based on video frames or snippets features \cite{Escorcia:2016:DAP, Buch:2017:SST, Buch:2017:SS-TAD, Yeung:2016:End, Gao:2017:TURN}.
In such way, actionness probabilities estimated at every time step can be analyzed and the potential start and end positions of the predicted action instances can be connected \cite{Zhao:2017:SSN, Lin:2018:BSN, Gong:2020:TSA, Yuan:2017:SMS, Lin:2020:Fast}.
In the second step, each generated proposal is classified into one of the action categories, whose boundaries could be refined by regression models \cite{Shou:2016:S-CNN, Gao:2017:CBR}.

\noindent
\textbf{Video Context Modeling.}
A most simple way to merge the features from multiple snippets is by directly applying pooling or convolution operations on the snippet features \cite{Shou:2016:S-CNN, Shou:2017:CDC}.
Also, multi-scale pooling or convolution has been widely employed to improve the feature representations \cite{Gao:2017:TURN, Zhao:2017:SSN, Yang:2020:TPN, Gong:2020:TSA}.
Besides the modeling by convolutions, some recent works have attempted to utilize graph convolutional networks (GCNs) or transformer \cite{attention2017} for long-term relationship modeling.
For example, G-TAD \cite{Xu:2020:G-TAD} embeds the temporal context with GCNs, while transformer based method RTDNet \cite{RTDNet:2021} is employed to directly generate proposals without post-process.
More recently, many methods \cite{Zeng:2019:P-GCN,localrich:2021,qtemporal:2021} pay attention to refine the proposals.
In particular, the video features and proposals generated by other proposal generation methods, \eg,  BSN\cite{Lin:2018:BSN} or BMN \cite{Lin:2019:BMN}, are all processed to enhance the proposal-level representations.
However, our proposal generation method only inputs video features and generates proposals by ours.

Apart from the above-mentioned ones, some other approaches may even exploit the context information at object-level \cite{Heilbron:2017:SSC, Wu:2020:CA-RCNN}.
However, those methods are beyond the scope of this paper since we mainly focus on the snippet-level feature learning and proposal generation while not proposal refinement.

\begin{figure*}[t]
\centering
\includegraphics[width=12cm,clip]{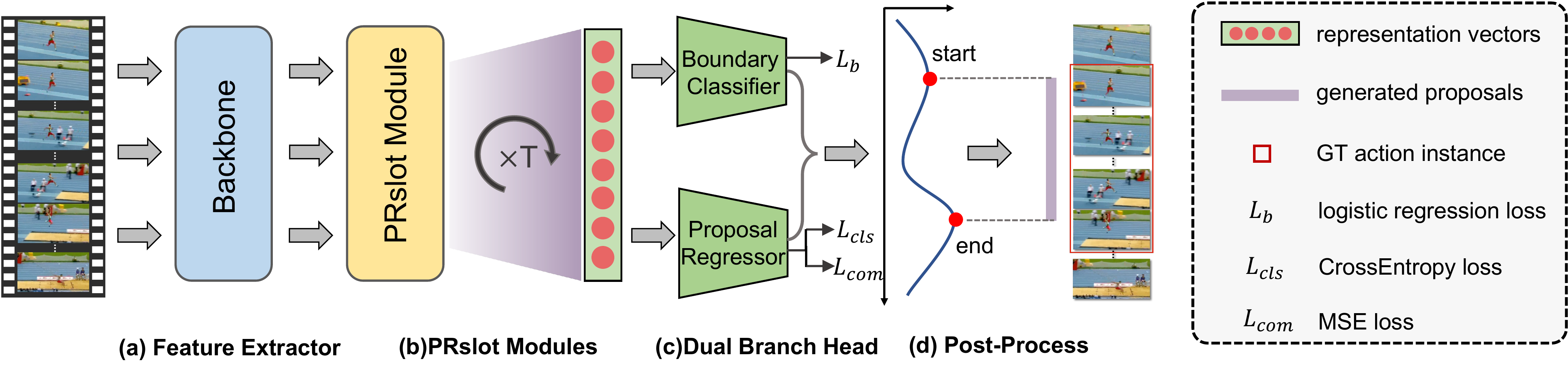}
\caption{The overall architecture of the proposed PRSA-Net.
The snippet-level features of videos are extracted by a CNN network in Feature Extractor. 
Then PRSlot modules are employed to contextualize slot representations with $T$ times. 
Next, our Boundary Classifier and Proposal Regressor are utilized to generate the predicted proposals.
Finally, post-process is utilized to  suppress redundant proposals.
}
\label{fig:arch}
\end{figure*}

\noindent
\textbf{Attention Mechanism.}
The vanilla attention mechanism is proposed by \cite{attention2017} and contributes more to capturing long-range dependencies.
Recently, attention mechanism has been widely applied in the computer vision field, \eg, image classification ViT \cite{vit2020}, video detection \cite{video2019}, group activity recognition \cite{li2021groupformer}, and object detection \cite{DETR2020}.
ViT \cite{vit2020} is the first work to employ a pure attention mechanism for image classification.
Original Slot Attention \cite{slotattn:2020} is proposed to exploit object-centric representation. 
In our work, different from the previous attention mechanism, our proposed region-based attention is estimated directly by high-level correlation features instead of similarity operations, which can concentrate on the local context better.
\section{Methodology}
\label{sec:method}
\subsection{Problem Formulation}
\label{sec:Form}
Assume that an arbitrary untrimmed video $V=\{ v_t \}_{t=1}^T$ contains a collection of $N$ action instances $\Psi = \{\psi_n = (t_{s,n}, t_{e,n}\}_{n=1}^{N}$, where $\psi_n$ refers to the $n$-th action instance and $(t_{s,n}, t_{e,n})$ correspond to its annotated start time, end time.
The aim of temporal action proposal generation task is to predict a set of action proposals $\Phi = \{\phi_w = (t_{s,w}, t_{e,w}, p_w\}_{w=1}^{W}$ as close to the ground truth annotations as possible based on the content of $V$. 
Here $p_w$ is the $w$-th proposal confidence.

\subsection{Overall Architecture}
\label{sec:arch}
We propose a Pyramid Region-based Slot Attention Network (PRSA-Net) to generate temporal action proposals precisely.
The pipeline of our PRSA-Net is illustrated in Fig. \ref{fig:arch}.
PRSA-Net consists of three major components: feature extractor, PRSlot modules and dual branch proposal generation head.
The original video is first fed into the feature extractor to produce the snippet-level features $X$ of size $L \times C_{\texttt{input}}$, where $L$ is the snippet length and $C_{\texttt{input}}$ is the feature dimension.
Then, PRSlot modules are employed to enhance action-matter contextual representations.
After this, a dual branch generation head maps the output embedding from the PRSlot modules to final boundaries and confidence scores respectively. 
The Boundary classifier is used to detect the boundaries,
while the proposal regressor is utilized to evaluate the proposal-level confidence.
We provide detailed descriptions of PRSA-Net below.

\subsection{Feature Extractor}
Given an untrimmed video $V$, we utilize the Kinetics \cite{Kay:2017:Kinetics} pre-trained I3D \cite{Carreira:2017:I3D} to extract video features.
We consider RGB and optical flow features jointly as they can capture different motion aspects.
Following the implementation of previous methods \cite{Lin:2018:BSN,Xu:2020:G-TAD}, each frame feature is flattened into a $C$-dimensional feature vector and then we group every consecutive $\sigma$ frames into $L = \ceil{T / \sigma}$ snippets.
In this work, these generated video snippets are the minimal units for further feature modeling.
For convenience, we denote the extracted snippet features as $X \in \mathbb{R}^{L \times C}$, where $C$ is the feature dimension and $L$ is the total number of video snippets.
In the end, a 1d convolution is applied to transform the channels into $C_{\texttt{input}}$.
For fair comparisons with \cite{Lin:2018:BSN,Lin:2019:BMN,Xu:2020:G-TAD}, we also conduct experiments based on the Kinetics pre-trained TSN \cite{Wang:2016:TSN} to extract video features.

\subsection{Pyramid Region-based Slot Attention}
\label{sec:CTR}
The original slot attention module is firstly proposed for updating the object-centric representations (\textit{slots}) by similarity-based attention.
Slots produced by slot attention bind to any object in the input embedding and are updated via a learned recurrent function for object-centric learning.
Inspired by this, we propose a tailor-modified Pyramid Region-based Slot Attention (PRSlot) module for capturing action-matter representations by a novel attention mechanism.

\subsubsection{\textbf{Overview.}}
We adopt dense slots to capture boundary-aware representation. 
The insight is that the motion content in boundaries context transfers rapidly and can be distinguished from the background.
The slot is initialized with snippet features $X$ and then is updated by the proposed PRSlot module for $T$ Times
For convenience, the $i$-th slot representation refined $t$ times denotes as $u^{(t)}_i$.
In particular, slot features are processed through convolutional layers with stride 1, window size 3 and padding 1, followed by a {ReLU} activation layer, and produce the slot features shaped as $L \times C_{\texttt{embed}}$.
Then, based on the proposed Region-based attention mechanism (\textit{will describe below}), we update our slot representations by the weighted sum of input and then apply {batchnorm} to the output vectors.

\begin{figure}[t]
  \centering
  \includegraphics[width=1.0\linewidth]{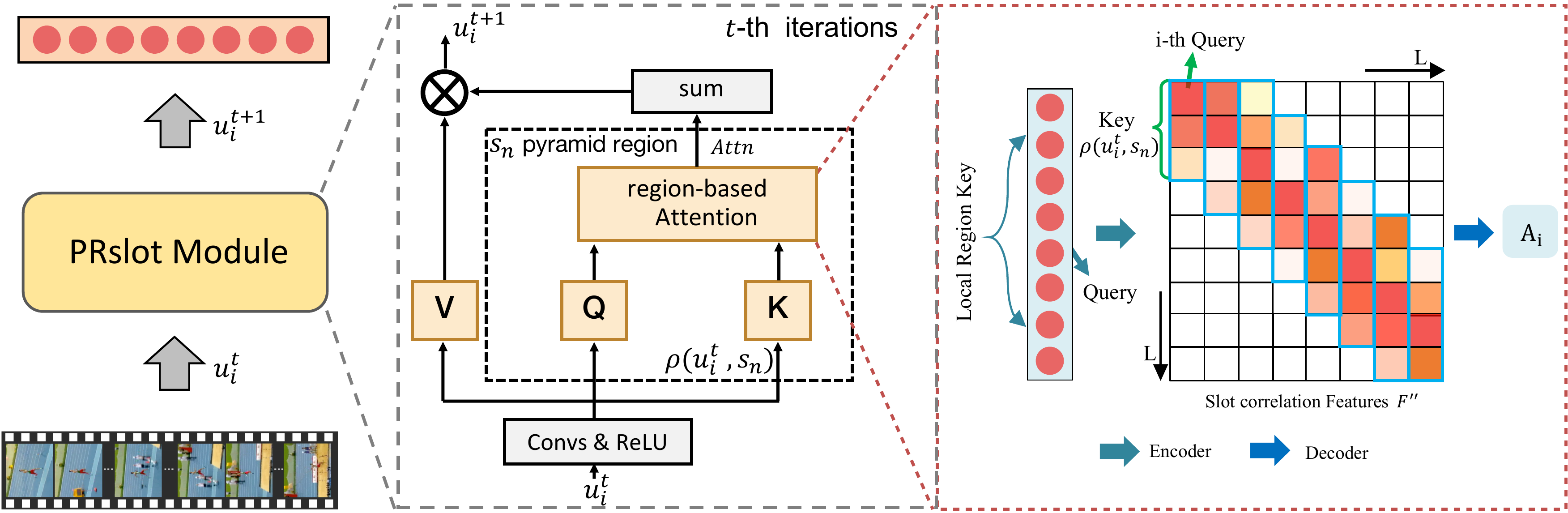}
  \caption{Overview of our Pyramid Region-based Slot Attention (PRSlot) module.
 For input snippet (slot) features $u_i$, we take $x_i$ as \textit{query} and its local context snippet (slots) feature as \textit{key}.
 An encoder-decoder operation is applied to map the correlation features into the local region score vector, and then generate the output slot features through the weighted sum of the input value (V).
  }
  \label{fig:MSSA}
\end{figure}

\subsubsection{\textbf{Region-based Attention Mechanism.}}
As shown in Fig. \ref{fig:MSSA}, to focus on the crucial local region relations, we design a newly region-based attention mechanism to learn the snippets relationships instead of using the similarity computation (\textit{cosine}, \textit{dot product} or \textit{Euclidean distance}).
Our region-based attention directly estimates the snippets interactions by the content of the snippet and its local surroundings.
Mathematically, given the target input feature $u_i$ and its surrounding features denoted as $\rho(u_i, s)= \{ u_{i-s}, ...,u_{i+s}\}$, where $s$ represents the window size of its local region, we formulate the calculation of the interaction scores as

\begin{equation}
  A_i = f_\mathrm{region}(u_i, \rho(u_i, s)).
  \label{eq:a-sum}
\end{equation}
Here the output $A_i$ is defined to be the attention vector for the $i$-th snippet (slot), and its length exactly equals $2s+1$, which is the temporal length of the covered surrounding $\rho(u_i, s)$.
$f_\mathrm{region}$ is a series of operations to map the local region features into the interaction confidence scores for the target snippet (slot). 
The value of $j$-th element in $A_i$, here denoted as $A_{i,j}$, is expected to indicate the relation score between $j$-th slot and the target slot $i$.
We take every input feature $u_i \in \{u_l\}_{l=0}^{L-1}$ as \textit{query}, and the local context slot of $u_i$ as \textit{key}, our designed region-based attention operation constructs a learnable score vector, which is used to quantitatively measure the importance between the target \textit{query} and \textit{key}.
In detail, this attention mechanism consists of an encoder to embed local contextual information and a decoder to estimate the relevant attention.
Due to the permutation invariant of temporal snippets, we additionally add the position embedding to the input features.
After this, the output slot representation can be described based on the attention over the feature map in the local region.

\noindent
\textbf{Encoder:}
Summarizing the surrounding slot context for the target slot is critical to augment slot representation.
The encoder is utilized to exploit contextual information for the surrounding slots of the target slot. 
The detailed implementation is described below.
Upon the input slot features $U \in \mathbb{R}^{C_{\texttt{embed}} \times L}$ (we omit the batch dimension for clarity), we repeat the temporal length $L$ in last dimension for sampling local windows, and then generate a 2D feature map $F \in \mathbb{R}^{C_{\texttt{embed}} \times L \times L}$.
A $1 \times 1$ convolutional layer is adopted to transform the $C_{\texttt{embed}}$-dimension channel into $C_{\texttt{out}}$-dimension channel.
Next, we apply the 2D convolution layer to augment the target slot using the local surrounding context shaped as $2s+1$.
Specifically, the 2D convolutional layer is implemented for target slot feature embedding, where we denote the kernel size as $(2s+1,1)$, padding size $(s,0)$ and stride $(1,1)$, indicating that we capture local context with size $2s+1$ from each column slot.
It generates the feature map $F' \in \mathbb{R}^{C_{\texttt{out}} \times L \times L}$, and we regard this feature map as the high-level correlation matrix.
For this content-based feature map $F'$, we only consider the local contextual information for each target slot.
The process can be formulated as,
\begin{equation}
F''_{i,j}=\left\{
\begin{array}{ccl}
F'_{i,j} \, , & &  \qquad {u_i \in \rho(u_j,s)},\\
0 \, , & & \qquad {\mathrm{Otherwise}},\\
\end{array} \right.
\end{equation}
where $F'_{i,j} \in \mathbb{R}^{C_{\texttt{embed}}}$ denotes the slot correlation features between $i$-th slot and $j$-th slot.

\noindent
\textbf{Decoder:}
For the sparse relation features $F''$, a decoder is deployed to map the relation features into the correspond relation scores.
It is noted that we only consider the local surroundings of the target slot and operate scores mapping, which can be formulated as,
\begin{equation}
A_{i,j}=\left\{
\begin{array}{ccl}
f_{\mathrm{region}}^{\mathrm{dec}}(F''_{i,j}) \, ,& & \qquad {u_i \in \rho(u_j,s)},\\
0  \, ,& & \qquad {\mathrm{Otherwise}},\\
\end{array} \right.
\end{equation}
Here $f_{\mathrm{region}}^{\mathrm{dec}}$ represents the decoding functions which is used to map the slots relation features into scores confidence.
In practice, a 2D convolutional layer is applied to distribute the relation features into relation scores, where we set kernel size to $(2s+1,1)$, padding size to $(s,0)$, out channel to $1$ and stride to $(1,1)$.
This convolutional layer is only deployed for the local slots $\rho(u_i,s)$ of target slot $i$ and formulate a slot confidence vector.
Finally, a $\mathrm{Softmax}$ operation is followed to normalize the score matrix.
In this way, each interaction score $A_{i,j}$ is actually dependent on the content of $u_j$ and $\rho(u_j,s)$.


\subsubsection{Iteration and update strategy.}
The original slot attention updates representations via recurrent function at each iteration $t=1...T$.
However, the recurrent function (\eg GRU) is time-consuming and achieves limited performance boost.
It will be demonstrated in ablation study.
In order to tackle the issues and make the PRSlot specialize in the temporal action-matter representation exploiting, we develop a parallel pyramid iteration strategy for slot representations updating.
Due to the variants of video duration, we apply a variety of local regions to exploit the slot representation completely. 
In detail, $|S|$ region-based attention using different scale surroundings $s \in S$ are deployed in parallel, where $S$ is a collection of snippet region size and $|S|$ denotes the carnality of the set.
In the end, we fuse the slot attention by element sum-wise and aggregate the input values to their assigned slots.

\subsection{Dual branch head}
Next, slot embedding provided by PRSlot modules serves as the latent action-aware representations for the proposal estimation.\\
\textbf{Boundary Classifier.}
A lightweight 1d convolution layer is introduced to transform the input channels into 2 for (\textit{start, end}) detection and a non-linear $\mathsf{sigmoid}$ function is followed to form the start/end probabilities $\mathcal{P}^s$/$\mathcal{P}^e$ separately.\\
\textbf{Proposal Confidence Regressor.}
Following the conventional proposal regression method \cite{Lin:2019:BMN}, we use pre-defined dense anchors to generate densely distributed proposals shaped as $D \times L$ and then apply 1DAlignLayer \cite{Xu:2020:G-TAD}
to extract the proposal-level features for corresponding proposal anchors shaped as $D \times L \times C_{\texttt{out}}$.
Finally, 2 FC layers are used to predict the proposal-level completeness maps $M^{com}$ and classification map $M^{cls}$.

\subsection{Training and Inference}
\label{sec:train}
\textbf{Label Assignment.}
We first generate temporal boundary label $G_s$ and $G_e$ followed by BSN \cite{Lin:2018:BSN}.
Next, we generate the dense proposal label map $G^c \in \mathbb{R}^{D \times L}$. 
As described in \cite{Lin:2019:BMN}, for a proposal $G^c_{i,j}$ with start frame $i$ and end frame $i+j$, we calculate the intersection over union (IoU) with all ground-truth and denote the maximum IoU as the value of $G^c_{i,j}$.\\
\textbf{Training.}
We define the following loss function to train our proposed model
\begin{equation}
  \mathcal{L}=\mathcal{L}_{{b}}+\mathcal{L}_{p}+ \lambda \mathcal{L}_{norm}
  \label{eq:loss}
\end{equation}
Here the boundary classification loss $\mathcal{L}_{b}$ is a weighted binary logistic regression loss used to determine the starting and ending scores $\mathcal{P}^s \in \{ \mathcal{P}_l^s \}_{l=1}^L$ or $\mathcal{P}^e \in \{ \mathcal{P}_l^e \}_{l=1}^L$.
$ \mathcal{L}_{norm}$ is the regularization term for network weights, we set $\lambda=0.0002$.
We also construct the proposal confidence losses $\mathcal{L}_{p}$ which is the combination of cross-entropy loss and mean square error loss to optimize dense proposals confidence scores $M^{com}$ and $M^{cls}$.
It can be formulated as,
\begin{equation}
    \mathcal{L}_p=\mathcal{L}_{cls}(M^{cls},G^c)+\lambda_{c} \mathcal{L}_{com}(M^{com},G^c)
\end{equation}
where $\mathcal{L}_{cls}$ is binary logistic regression loss and  $\mathcal{L}_{com}$ is the MSE loss, usually we set the balance term $\lambda_c=10$.\\
\textbf{Inference.}
Based on the outputs of the boundaries classifier, we select the valid starting snippets from $\{\mathcal{P}^s_{l}\}_{l=1}^L$ by two conditions:
(1) $\mathcal{P}^{s}_{l-1}<\mathcal{P}^{s}_{l}; \mathcal{P}^{s}_{l}>\mathcal{P}^{s}_{l+1}$;
(2) $\mathcal{P}^s_{l}>0.5 \times \max_{n=1}^L\{\mathcal{P}_n^s\}$.
We apply the same rule for recording ending snippets.
Then, we combine these valid starting and ending snippets and obtain the candidate proposals denoted as $\Phi = \{\phi_w = (t_{s,w}, t_{e,w}, p_w, M_w \}_{w=1}^{W}\}$, where $\mathcal{P}_w=\mathcal{P}^{s}_{t_{s,w}} \cdot \mathcal{P}^{e}_{t_{s,w}}$ and proposal-level confidence $M_w=M^{cls}_{t_{s,w},t_{e,w}} \cdot M^{com}_{t_{s,w},t_{e,w}}$.
Finally, we fuse the proposal confidence scores and output the predicted proposals $\Phi = \{\phi_w = (t_{s,w}, t_{e,w}, S_w \}_{w=1}^{W}\}$, here $S_w=p_w \times M_w$.\\
\textbf{Post-Processing.}
The non-maximum suppression (NMS) algorithm \cite{neubeck:2006:NMS} is adopted to suppress redundant proposals with high overlaps.
Also, we use the Soft-NMS \cite{softnms2017} algorithm to suppress the proposals and report the performance for a fair comparison with previous methods.

\section{Experiments}
\label{sec:expm}

\subsection{Datasets and Evaluation Metrics}
\textbf{THUMOS14 \cite{Idrees:2017:THUMOS}.} THUMOS14 dataset contains respectively 200 videos in the validation for training and 213 videos in the test set for inference.
It has a total number of 20 classes, and each video has around 15 action instances on average.\\
\textbf{ActivityNet-1.3 \cite{Caba:2015:ActivityNet}.}
ActivityNet-1.3 dataset contains 19994 untrimmed videos labeled in a wider range of 200 action categories with a lower 1.5 action instances per video on average.
These videos are split into the training, validation, and test set by the ratio of 2:1:1.
We evaluate on the validation set of ActivityNet-1.3.\\
\textbf{Evaluation metric.}
To assess the performances of action proposal generation, we report the average recall (AR) under different intersections over union thresholds (tIoUs) with various average number of proposals (AN) for each video.
Following the conventions, we adopt the tIoUs of $\{0.5:0.05:1.0\}$ on THUMOS14 and $\{0.5:0.05:0.95\}$ on ActivityNet-1.3.
To evaluate the quality of our generated proposals, we also evaluate the performances of action detection using the mean average precision (mAP) at different tIoUs.
Following the official evaluation API, the tIoUs of $\{0.3, 0.4, 0.5, 0.6, 0.7\}$ are used for THUMOS14, and $\{0.5, 0.75, 0.95\}$ are used for ActivityNet-1.3.

\subsection{Experimental Setup}
Following the conventional setting, we extracted the 2048-dimensional video features by two-stream I3D \cite{Carreira:2017:I3D} pre-trained on Kinetics \cite{Kay:2017:Kinetics} on THUMOS14.
Besides, for a fair comparison, we also conduct experiments based on the two-stream network TSN \cite{Wang:2016:TSN} backbone, where ResNet network \cite{resnet2016} and BN-Inception network \cite{inception2015} are applied as the spatial and temporal network respectively.
We also set $C_{\texttt{input}}=256$. 
On ActivityNet-1.3, the video features were extracted by the pre-trained model provided in \cite{Xiong:2016:CUHK, Caba:2015:ActivityNet}.
In data preparation, we set the snippet interval $\sigma$ to 4 on THUMOS14 and 16 on ActivityNet-1.3.
Then we cropped each video feature sequence with the overlapped windows of stride of $100$ and length $L = 250$ on THUMOS14.
While on ActivityNet-1.3, we set the temporal length to $L = 100$ by linear interpolation.
Also, we enumerate all possible anchors where the max durations is $D=64$ on THUMOS14 and $D=100$ on ActivityNet-1.3.
For the PRSlot module, we set the embed dimension $C_{\texttt{embed}}=256$, $C_{\texttt{out}}=256$, and the local region scale $S=\{4, 8\}$.
In post-processing, we set the NMS threshold $\theta$ to 0.65 and 0.45 on THUMOS14 and ActivityNet-1.3 respectively.
During the training, it was optimized on one NVIDIA TESLA V100 GPU with batchsize 16 on ActivityNet1.3 and 8 on THUMOS14.
We adopted the Adam optimizer \cite{Kingma:2019:Adam} for network optimization.
For THUMOS14, the models were tuned for 10 epochs with the learning rate set to $2 \times 10^{-4}$.
For ActivityNet-1.3, we trained our models by setting the learning rate to $10^{-3}$ in the first 7 epochs and decaying it to $10^{-4}$ in the last 3 epochs.
\begin{table}[t]
\centering
\caption{Comparison of the action proposal generation performances with state-of-the-arts on THUMOS14 in terms of AR@AN(\%).}
\scalebox{0.7}{
\begin{tabular}{p{4.0cm}p{2.0cm}<{\centering}p{1.3cm}<{\centering}p{1.3cm}<{\centering}p{1.3cm}<{\centering}p{1.3cm}<{\centering}}

\toprule
Method &Backbone & @50 & @100 & @200  & @500\\
\midrule
MGG \cite{Liu:2019:MGG}         &TSN   & 39.93 & 47.75 & 54.65 & 61.36\\
BSN \cite{Lin:2018:BSN} + SNMS  &TSN     & 37.46 & 46.06 & 53.21 & 60.64\\
BMN \cite{Lin:2019:BMN} + SNMS  &TSN     & 39.36 & 47.72 & 54.70 & 62.07\\
BC-GNN \cite{BCGNN2020} + NMS &TSN  & 41.15& 50.35 & 56.23 & 61.45\\
BU-TAL \cite{Zhao:2020:Bottom} &I3D & 44.23 & 50.67 & 55.74 & -\\
BSN++ \cite{bsn++2020} +SNMS &TSN &42.44&49.84&57.61&65.17\\
RTD-Net \cite{RTDNet:2021} &I3D &41.52 &49.32 &56.41 &62.91 \\
\hline
Ours + NMS &TSN &{47.49} & {55.14} & {60.18} & 63.53\\
Ours + SNMS &TSN & 44.11 & 52.52 & 59.19 & {65.12}\\
Ours + NMS &I3D & \textbf{49.06} & \textbf{56.12} & \textbf{61.30} & 63.20\\
Ours + SNMS &I3D &45.81  &53.13  &59.32  &\textbf{66.32} \\
\bottomrule
\end{tabular}
}

\label{tab:THUMOS14-proposal}
\end{table}

\subsection{Temporal Proposal Generation}
\begin{table}[t]
\vspace{-0.1cm}
\centering
\caption{Comparison of the action proposal generation performances with state-of-the-arts on ActivityNet-1.3 in terms of AR@AN(\%) and AUC.}
\scalebox{0.62}{
\begin{tabular}{p{2.2cm}<{\centering}p{2.2cm}<{\centering}p{2.2cm}<{\centering}p{2.2cm}<{\centering}p{2.2cm}<{\centering}p{2.2cm}<{\centering}p{2.2cm}<{\centering}p{2.2cm}<{\centering}}

\toprule
Metric &
BSN \cite{Lin:2018:BSN} \quad &
MGG \cite{Liu:2019:MGG}  \quad &
BMN \cite{Lin:2019:BMN} \quad  &
BC-GNN \cite{BCGNN2020} \quad &
BU-TAL \cite{Zhao:2020:Bottom} \quad &
RTD-Net \cite{RTDNet:2021} \quad &
Ours\\
\midrule
AR@1   & 32.17 &  -   & -    &  -   & -   & 33.05   &\textbf{35.37} \\
AR@100 & 74.16 & 74.54 & 75.01&76.73 & 75.27 &73.21& \textbf{76.90}\\
AUC    & 66.17 & 66.43 & 67.10&68.05 & 66.51 &65.78& \textbf{69.21}\\
\bottomrule
\end{tabular}
}
\vspace{-0.3cm}
\label{tab:acnet-proposal}
\end{table}

\subsubsection{Comparisons with State-of-the-Arts.}
We compare our PRSA-Net with other state-of-the-art methods on two backbones for a fair comparison.
The results on THUMOS14 are summarized in Table \ref{tab:THUMOS14-proposal}.
It can be observed that our proposed PRSA-Net outperforms all of the aforementioned methods by a large margin with either NMS or Soft-NMS used in post-processing.
For instance, when using TSN as feature extractor, our method respectively improve the AR@50 from 42.44\% to 47.49\%, AR@100 from 49.84\% to 55.14\%, and AR@200 from 57.61\% to 60.18\%.
It is worth noting that our model with I3D backbone outperforms all previous methods by a large margin (+ 4.83\% AR@50 and +5.45\% AR@100).
With the well-design PRSlot and its region-based attention, our method establishes new state of the art performance on THUMOS14.
The results on ActivityNet-1.3 are displayed in Table \ref{tab:acnet-proposal}.
Our method shows better performances than the previous best results under both AR@100 and AUC scores.
In particular, the AR@1 achieves 2.32\% performance boost in this well-studied benchmark.

\subsection{Ablation Studies}
\label{sec:ablation}

We conduct extensive experiments on THUMOUS14 using the I3D \cite{Wang:2016:TSN} backbone and NMS for post-processing to investigate the effectiveness and proper settings of the different components of our proposed models.\\
\begin{table}[t]
\centering
\caption{Comparisons of different combinations in our model. Evaluated on THUMOS14 in terms of AR@AN(\%). \texttt{SA} and \texttt{RA} denote the implementation of similarity-based and our region-based attention separately. \texttt{original} and \texttt{ours} denote the original and our parallel pyramid strategy respectively.}
\scalebox{0.8}{
\begin{tabular}{p{2.0cm}<{\centering}p{2.0cm}<{\centering}|p{2.0cm}<{\centering}p{2.0cm}<{\centering}|p{1.2cm}<{\centering}p{1.2cm}<{\centering}p{1.2cm}<{\centering}}
\hline
\multicolumn{2}{c|}{{Attention Mechanism}} &\multicolumn{2}{c|}{{Iteration Strategy}}& \multicolumn{3}{c}{AR@AN (\textit{testing set)}} \\

$\texttt{SA}$ & $\texttt{RA}$ & \texttt{original}&\texttt{ours}&@50 & @100 & @200  \\

\hline
\cmark &    & \cmark &  & 35.6  &43.1  &50.9 \\
\cmark   &  &  &\cmark  & 43.5 & 52.3 & 56.5\\
   & \cmark  & \cmark& & 42.9 & 52.8 & 55.6\\
 & \cmark & &\cmark &\textbf{49.1}&\textbf{56.1}&\textbf{61.3}\\
\bottomrule
\end{tabular}
}

\label{tab:clue-combinations}
\end{table}

\begin{table}[t]
\centering
\caption{Comparisons of different local region scale choices in region-based attention. Evaluated on THUMOS14 in terms of AR@AN(\%).}
\scalebox{0.8}{
\begin{tabular}{p{4cm}<{\centering}p{1.7cm}<{\centering}p{1.7cm}<{\centering}p{1.7cm}<{\centering}p{1.7cm}<{\centering}}
\toprule
Local Region Scale&@50 &@100 &@200 &@500 \\
\midrule

2 &47.2 &54.3 &59.4 &62.9\\

4 &47.8 &55.4 &59.9 &63.2 \\

8 &48.8 &56.0 &60.6 &63.0\\

12 &49.0&56.1 &60.5 &62.7\\
\hline

2, 4 &48.6 &55.7 &60.4 &63.0\\
4, 8 &\textbf{49.1} &\textbf{56.1} &61.3 &63.2\\
2, 4, 8 &{49.1} &{56.1} &60.2 &63.2\\

4, 8, 12 &48.7&55.8&\textbf{61.5}&\textbf{63.4}\\
\bottomrule
\end{tabular}
}

\label{tab:scale-choice}
\end{table}

\vspace{-0.3cm}

\begin{table}[t]
\centering
\caption{Different setting choices for the number of pyramid iterations.
Evaluated on THUMOS14 in terms of AR@AN (\%)}
\scalebox{0.7}{
\begin{tabular}{p{4cm}<{\centering}p{2cm}<{\centering}p{2cm}<{\centering}p{2cm}<{\centering}p{2cm}<{\centering}}
\toprule
Iteration times & @50 &@100&@200&@500 \\
\midrule
1 &48.4&55.3&60.1&61.6                      \\
2 &\textbf{49.1}&\textbf{56.1}&61.3&\textbf{63.2}            \\
3 &48.9&55.7&\textbf{61.9}&62.3    \\
\bottomrule
\end{tabular}
}

\label{tab:blocks}
\end{table}
\begin{figure}[t]
  \centering
  \includegraphics[width=0.8\linewidth]{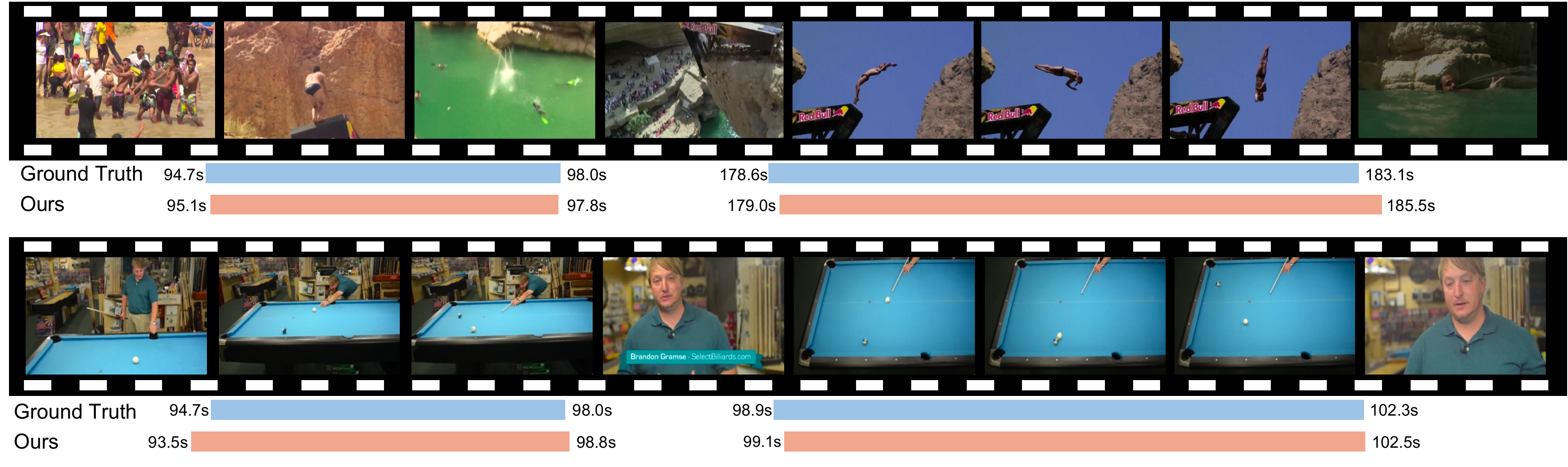}
  \caption{Some action proposal examples predicted by our PRSA-Net on THUMOS14.}
  \label{fig:quality_result}
  \vspace{-0.1cm}
 \end{figure}
 
\noindent 
\textbf{Variants of our PRSA-Net.}
To measure the importance of our proposed region-based attention and parallel pyramid iteration strategy, we conduct the ablation study with the following variants.
Baseline: we replace the PRSlot with the original slot attention, which includes the similarity-based attention and recurrent iteration strategy.
More baseline details can be found in our Supplementary materials.
We also adopt different combinations of our PRSlot architecture components.
Table \ref{tab:clue-combinations} reports the detailed proposal generation results on THUMOS14.
The \cmark \, symbols stand for \textit{with} the corresponding components or strategies.
We can find that our proposed region-based attention (\textit{the last row}) improves the AR by more than 6\%, indicating the effectiveness of our designed PRSlot module.
Additionally, our parallel pyramid iteration strategy also contributes to performance boosts.
We attribute performance differences to variations in architecture design and iteration schemes.

\noindent
\textbf{Scale of local region.}
Next, we investigate the effect of the local region size.
The results on THUMOS14 are displayed in Table \ref{tab:scale-choice}.
The first column of the table lists the corresponding choices of $S$ in a parallel pyramid strategy.
We can find that (1) a single scale with large $s = 12$ can perform already quite well; and
(2) the multi-scale regions of $\{4, 8\}$ and $\{4, 8, 12\}$ alternatively reach the best results under different AN.
These results demonstrate the effectiveness of our parallel pyramid strategy for complementary context representation modeling.
To balance the performance and model simplicity, we implement our models with $\{4, 8\}$.\\
\textbf{Sensitivity to iteration times.}
We also report the results of the sensitivity of our model to the different settings for iteration times in Table \ref{tab:blocks}.
We find that using $2$ iterations can reach the best results while performance slightly degrades at using $3$ times.

%

\subsection{Qualitative Results}
In Fig. \ref{fig:quality_result}, we visualize some action detection results generated by our PRSA-Net on THUMOS14.
We can find that our model can successfully detect the action instances in these examples with precise action boundaries.
For the untrimmed video with multiple action instances, the foreground actions and background scenes can be well distinguished.

\subsection{Action Detection with our proposals}

\begin{table}[t]
\small
\centering
\caption{Performance comparisons of temporal action detection on THUMOS14 in terms of mAP@tIoUs(\%).}
\scalebox{0.7}{
\begin{tabular}{p{3cm}<{\centering}p{2cm}<{\centering}p{2cm}<{\centering}p{1.5cm}<{\centering}p{1.5cm}<{\centering}p{1.5cm}<{\centering}p{1.5cm}<{\centering}p{1.5cm}<{\centering}}
\toprule
Method &Backbone &Classifier & 0.7  & 0.6 & 0.5  &0.4  &0.3\\
\midrule
BSN \cite{Lin:2018:BSN}          &TSN  &UNet &20.0 &28.4 &36.9 &45.0 &53.5\\
MGG \cite{Liu:2019:MGG}          &TSN  &UNet &21.3 &29.5 &37.4 &46.8 &53.9\\
BMN \cite{Lin:2019:BMN}          &TSN  &UNet &20.5 &29.7 &38.8 &47.4 &56.0\\
G-TAD \cite{Xu:2020:G-TAD}       &TSN  &UNet &23.4 &30.8 &40.2 &47.6 &54.5\\
BU-TAL et al. \cite{Zhao:2020:Bottom} &I3D &UNet &28.5 &38.0 &45.4 &50.7 &53.9\\
BSN++ \cite{bsn++2020}           &TSN  &UNet &22.8&31.9&41.3&49.5&59.9\\
BC-GNN \cite{BCGNN2020}          &TSN  &UNet &23.1&31.2 &40.4  &49.1 &57.1\\
RTD-Net \cite{RTDNet:2021}               &I3D  &UNet &25.0 &36.4 &45.1 &53.1  &58.5   \\
\hline
Ours &TSN  &UNet &{28.8} &{39.2} &{51.1} &{58.9} &{65.4}\\
Ours &I3D  &UNet &\textbf{30.9} &\textbf{44.0} &\textbf{55.0} &\textbf{64.4} &\textbf{69.1}\\

\hline
BSN \cite{Lin:2018:BSN}          &TSN  &PGCN &- &- &49.1 &57.8 &63.6\\
G-TAD \cite{Xu:2020:G-TAD}       &TSN  &PGCN &22.9 &37.6 &51.6 &60.4 &66.4 \\
RTD-Net \cite{RTDNet:2021}               &I3D  &PGCN &23.7 &38.8 &51.9 &62.3  &68.3   \\
\hline
Ours                            &I3D   &PGCN  &\textbf{28.4}&\textbf{47.3}&\textbf{58.7}&\textbf{73.2}&\textbf{76.3} \\
\bottomrule
\end{tabular}
}
\label{tab:THUMOS14-localization}
\end{table}

\begin{table}[t]
\small
\centering
\caption{Comparison with state-of-the-arts detection methods on ActivityNet-1.3 in terms of mAP@tIoUs(\%).}
\scalebox{0.8}{
\begin{tabular}{p{2cm}<{\centering}p{2cm}<{\centering}p{2cm}<{\centering}p{2cm}<{\centering}p{2cm}<{\centering}}
\toprule
Method & 0.5 & 0.75 & 0.95 & Average\\
\midrule
TAL-Net \cite{Chao:2018:TAL-Net} &38.23 &18.30 &1.30 &20.22\\
BSN \cite{Lin:2018:BSN}          &46.45 &29.96 &8.02 &30.03\\
P-GCN \cite{Zeng:2019:P-GCN}     &48.26 &33.16 &3.27 &31.11\\
BMN \cite{Lin:2019:BMN}          &50.07 &34.78 &8.29 &33.85\\
G-TAD \cite{Xu:2020:G-TAD}       &50.36 &34.60 &9.02 &34.09\\
BSN++ \cite{bsn++2020}           &51.27 &35.70 &8.33    &34.88\\
BC-GNN \cite{BCGNN2020}          &50.56 &35.35 &{9.71}    &34.68\\
RTD-Net \cite{RTDNet:2021}       &47.21 &30.68 &8.61 &30.83 \\
\hline
Ours &\textbf{52.37} &\textbf{37.18} &\textbf{9.78} &\textbf{36.26}\\
\bottomrule

\end{tabular}
}
\vspace{-0.5cm}
\label{tab:acnet-localization}
\end{table}

When evaluating the quality of our proposals, we follow the conventional two-stage action detection works \cite{Xu:2020:G-TAD,Lin:2019:BMN,Lin:2018:BSN}.
Therefore, we classify the candidate proposals using external classifiers.
We use the video classifier in UntrimmedNet \cite{Wang:2017:Unet} to assign the video-level action classes on THUMOS14, while on ActivityNet1.3, we adopt the video-level classification results from \cite{Xiong:2016:CUHK} to assign the action labels to detect the action class.
Furthermore, we also introduce the proposal-level classifier P-GCN \cite{Zeng:2019:P-GCN} to predict action labels for every candidate proposals.
We evaluate the final action detection performances and make comparisons with the state-of-the-art.
The results on THUMOS14 are summarized in Table \ref{tab:THUMOS14-localization}.
Compared with the other state-of-the-art methods, our approach achieves significant improvements under all tIoU settings.
Especially, the mAP at the typical IoU = 0.5 was boosted from 45.4\% to 55.0\%, reaching a considerable improvement ratio of 21.1\%.
Also, when proposal-level classifier P-GCN is applied to classify our generated proposals following the same implementations in \cite{Xu:2020:G-TAD,RTDNet:2021}, the performance can be boosted rapidly and achieve 58.7\% mAP@0.5.
Table \ref{tab:acnet-localization} shows our action detection results on the validation set of ActivityNet-1.3 compared with previous works.
Again, our method outperforms most other methods by a large margin in almost all cases, including the average mAPs over different tIoUs.
These experiments demonstrate that the proposals generated by our PRSA-Net are able to boost the action detection performance better.



\section{Conclusion}
\label{sec:end}

In this paper, we propose a novel Pyramid Region-based Slot Attention Network (PRSA-Net) for temporal action proposal generation.
Specifically, a Pyramid Region-based Slot Attention (PRSlot) module is introduced to capture action-aware representations, which is enhanced by the proposed region-based attention and parallel pyramid iteration strategy.
The experiments show the advantages of our method both in action proposals and action detection performance.

%
%
\bibliographystyle{splncs04}
\bibliography{egbib}

\end{document}